\documentclass[accepted]{uai2026} % for initial submission
\usepackage[american]{babel}
\usepackage{natbib} % has a nice set of citation styles and commands
\usepackage{mathtools} % amsmath with fixes and additions
\usepackage{booktabs} % commands to create good-looking tables
\usepackage{tikz} % nice language for creating drawings and diagrams
\usepackage{multirow}
\usepackage{amssymb}

\usepackage[ruled,vlined]{algorithm2e}

\newtheorem{remark}{Remark}

\title{ Function-Space Empirical Bayes Regularisation with Student's t Priors }

\author[1]{\href{mailto:<pengchenghao@sz.tsinghua.edu.cn>?Subject=Your UAI 2026 paper}{Pengcheng~Hao}{}}
\author[1]{\href{mailto:<kuruoglu@sz.tsinghua.edu.cn>?Subject=Your UAI 2026 paper}{Ercan~Engin~Kuruoglu}{}}
\affil[1]{%
    Institute of Data and Information\\
    Tsinghua Shenzhen International Graduate School\\
    Shenzhen, China
}

\begin{document}
\maketitle

\begin{abstract}
Bayesian deep learning (BDL) has emerged as a principled approach to produce reliable uncertainty estimates by integrating deep neural networks with Bayesian inference, and the selection of informative prior distributions remains a significant challenge. Various function-space variational inference (FSVI) regularisation methods have been presented, assigning meaningful priors over model predictions. However, these methods typically rely on a Gaussian prior, which fails to capture the heavy-tailed statistical characteristics inherent in neural network outputs.
By contrast, this work proposes a novel function-space empirical Bayes regularisation framework---termed ST-FS-EB---which employs heavy-tailed Student's $t$ priors in both parameter and function spaces. Also, we approximate the posterior distribution through variational inference (VI), inducing an evidence lower bound (ELBO) objective based on Monte Carlo (MC) dropout. Furthermore, the proposed method is evaluated against various VI-based BDL baselines, and the results demonstrate its robust performance in in-distribution prediction, out-of-distribution (OOD) detection and handling distribution shifts.
\end{abstract}

\section{Introduction}\label{sec:intro}
% Deep neural network cannot qualify uncertainty % weight-space BNN training lacks informative prior and precise posterior
The remarkable predictive accuracy of deep neural networks (DNNs) has established them as the cornerstone of modern artificial intelligence. Yet, a closer examination reveals a fundamental blind spot: a DNN, no matter how accurately it predicts, remains oblivious to what it does not know~\cite{gawlikowski2023survey}. This distinction between "knowing" and "knowing that one knows" lies at the heart of intelligent decision-making. In high-stakes environments, a model's silence about its own uncertainty—e.g., failing to flag misdiagnosis risks in medical imaging~\cite{lambert2024trustworthy} or misjudging pedestrian presence in autonomous driving~\cite{wang2025uncertainty}—represents not merely a technical limitation but a practical hazard.
The question then is not only how to make predictions but also how to equip models with the capacity to express doubt. Bayesian neural networks (BNNs) offer a principled response to this challenge~\cite{jospin2022hands}. By casting network weights as random variables governed by prior distributions and pursuing posterior inference, BNNs move beyond point estimates to produce predictions together with uncertainty. To infer the posterior distribution over weights, researchers have explored a variety of approaches, including Markov Chain Monte Carlo (MCMC), Laplace approximation, and variational inference (VI). Most BNNs employ Gaussian priors, yet both network parameters and predictive distributions often exhibit heavy-tailed behaviour in practice~\cite{fortuin2022bayesian}. 
To accommodate this heavy-tailed behaviour, recent works have adopted heavy-tailed distributions as priors for the model weights~\cite{fortuin2021bnnpriors,xiao2023heavy,harrison2025heteroscedastic}.
However, these weight-space methods struggle to incorporate meaningful prior knowledge into BNNs, as the relationship between weights and predictions is highly intricate.

% function-space regularisation and its drawbacks
An alternative paradigm shifts inference from weight space to function space. Instead of placing priors on model parameters, functional VI (FVI)~\cite{sun2018functional} defines interpretable prior distributions directly over model predictions and optimises a functional evidence lower bound (ELBO). However, this conceptual shift comes at a cost: the Kullback–Leibler (KL) divergence between infinite-dimensional stochastic processes lacks a closed-form expression.~\cite{sun2018functional} approximates the KL divergence with the spectral Stein gradient estimator (SSGE)~\cite{shi2018spectral}, introducing significant computational overhead and limiting the scalability of the FVI in real-world settings. By contrast, linearisation-based function-space VI (FSVI)~\cite{rudner2022tractable} methods impose Gaussian priors on model parameters and approximate the induced functional distribution as a Gaussian process (GP). This yields a tractable, closed-form approximation of the KL divergence between prior and posterior~\cite{immer2021improving, rudner2022tractable, cinquin2024regularized}. However, the linearisation strategy, though convenient, distorts the true function mapping and imposes a heavy computational burden due to the need to evaluate Jacobian matrices. By contrast, the function-space empirical Bayes (FS-EB) approach~\cite{rudner2023functionspace, rudner2024finetuning} introduces regularisation simultaneously in both parameter and function spaces, circumventing the need for linearisation. However, all these functional methods are built upon Gaussian assumptions, limiting their ability to adequately model heavy-tailed behaviour.

%Our proposed method
In comparison, this work considers a new FS-EB framework based on Student's $t$ priors  (ST-FS-EB). The benefits are twofold: 1) the heavy-tailed nature of the Student’s $t$ distribution provides enhanced robustness to outliers and model misspecification compared to the Gaussian distribution; 2) the additional degree-of-freedom (dof) parameter offers greater flexibility in capturing the tail decay of the true underlying distribution. The main contributions of this work are shown below:
\begin{enumerate}
\item We present a Student's $t$ prior-based FS-EB framework, which employs an empirical functional prior derived from Student's $t$-distributed weights and a functional likelihood defined by a Student's $t$ process.
\item We adopt Monte Carlo (MC) dropout as the variational distribution to approximate the posterior within a VI framework, deriving an MC dropout-based ELBO objective.
\item  Our method is compared against a range of function-space and parameter-space regularisation baselines on five real-world benchmarks. The results demonstrate that ST-FS-EB achieves robust performance under both standard and distribution-shift settings.
\end{enumerate}
The remainder of this paper is organised as follows. Section~\ref{sec: RW} reviews the related work, while Section~\ref{sec: pre} presents the theoretical background. Section~\ref{sec: proposed} then introduces the proposed ST-FS-EB method. The experimental evaluation is reported in Section~\ref{sec: simulation}, and Section~\ref{sec: conclusion} concludes the paper.

\section{Related Work}\label{sec: RW}
\subsection{Weight-space Variational Inference Regularisation}

\cite{hinton1993keeping} pioneers the first variational Bayesian approach for neural networks, using a diagonal Gaussian approximation of the weight posterior to implement minimum description length (MDL) regularisation. Also,~\cite{graves2011practical} extends the variational method to complex neural network architectures by estimating the gradients of model parameters via MC sampling. However, both methods require substantial computational costs, making them impractical for large-scale neural networks. By contrast,~\cite{kingma2014auto}  introduces a reparameterization trick, which enables straightforward optimisation of the variational ELBO using standard stochastic gradient methods. Then~\cite{blundell2015weight} introduces a Bayes by Backprop method, a VI algorithm compatible with backpropagation that learns distributions over model weights by minimising the ELBO. Building upon this line of research, \cite{gal2016dropout} proposes the MC dropout method, establishing a theoretical connection between dropout training and approximate Bayesian inference. 

\subsection{Function-space Variational Inference Regularisation}\label{sec: FSVI_R}
Initial FSVI studies focus on directly approximating posterior distributions over functions induced by neural networks. In particular,~\cite{wang2018function} proposes a function-space particle optimisation framework, performing VI over regression functions. Also,~\cite{flam2017mapping} presents a function-space prior alignment approach that maps GP priors to BNN priors, allowing BNNs to inherit structured functional properties from GPs. Additionally,~\citet{sun2018functional} demonstrates that the KL divergence between infinite-dimensional prior and variational functional processes can be expressed as the supremum of marginal KL divergences over all finite input sets, which allows the functional ELBO to be approximated using finite measurement sets. However, the employed SSGE method requires high computational cost, and the function-space variational objective is generally ill-defined~\cite{burt2021understanding}. By contrast, an efficient VI framework for posterior approximation is enabled by \cite{ma2019variational} through the use of functional implicit stochastic process priors to handle intractable function-space objectives. Also, a well-defined variational objective is achieved in~\cite{ma2021functional} through a grid-functional KL divergence built on stochastic process generators. In comparison,~\cite{khan2019approximate,immer2021improving} formulate a VI objective by modelling DNNs as GPs using Laplace and GGN approximations. To enable variance parameter optimisation, \cite{rudner2022tractable} further introduces a fully BNN-based FSVI framework, in which the KL divergence is rendered finite by defining the prior as the pushforward of a Gaussian distribution in weight space. Moreover, \cite{cinquin2024regularized} proposes an interpretable prior construction based on a pretrained GP prior and introduces a regularised KL divergence to ensure a well-defined variational objective. However, these approaches approximate the model output distribution through linearisation, which requires computationally expensive Jacobian evaluations and simultaneously introduces additional approximation error.
In comparison, the FS-EB approach~\cite{rudner2023functionspace,rudner2024fine,rudner2024mind} avoids linearisation and constructs an empirical functional prior by combining a weight-space prior with a GP-based functional likelihood.

\section{Preliminary}\label{sec: pre}
This section presents the fundamental concepts that form the foundation of our proposed ST-FS-EB framework. Specifically, Section~\ref{sec:ST distribution} and Section~\ref{sec:ST process} introduce the Student’s $t$ distribution and the Student’s $t$ process, respectively. Furthermore, Section~\ref{sec: FSEB_method} describes the FS-EB framework, which serves as the basis for our method.

\subsection{The Student's t distribution}\label{sec:ST distribution}
The Student's $t$ distribution~\cite{ahsanullah2014normal} is a continuous probability distribution, which can be viewed as a heavy-tailed generalisation of the Gaussian distribution and provides increased robustness to outliers. A univariate Student's $t$ random variable $x \in \mathbb{R}$ with location parameter $\mu \in \mathbb{R}$, scale parameter $\sigma > 0$ and dof $\nu > 0$ is denoted by $x \sim \mathcal{ST}(\nu, \mu, \sigma^2)$. 
Its probability density function (PDF) is given by 
\begin{equation}
p(x) = \frac{\Gamma\left(\frac{\nu+1}{2}\right)}{\Gamma\left(\frac{\nu}{2}\right) \sqrt{\pi\nu\sigma^2} } \left(1 + \frac{(x-\mu)^2}{\nu\sigma^2}\right)^{-\frac{\nu+1}{2}}, \label{eq:student_t_pdf} 
\end{equation} 
where $\Gamma(\cdot)$ denotes the Gamma function. Compared with the Gaussian distribution, the polynomial decay of the tails in \eqref{eq:student_t_pdf} assigns significantly higher probability mass to extreme events. Smaller values of $\nu$ correspond to heavier tails, implying higher robustness to outliers. Also, the Student's $t$ distribution includes the Gaussian distribution as a limiting special case. Specifically, as the dof parameter $\nu \to \infty$, the Student's $t$ distribution converges to a normal distribution: 
\begin{equation*} 
\mathcal{ST}(\nu, \mu, \sigma^2) \xrightarrow[\nu \to \infty]{} \mathcal{N}(\mu, \sigma^2). 
\end{equation*} 
This property establishes a smooth continuum between heavy-tailed and Gaussian modelling, allowing the Student's $t$ distribution to adaptively interpolate depending on the dof value $\nu$. Besides, assume a random vector $\mathbf{x} \in \mathbb{R}^d$ follows a multivariate Student's $t$ distribution~\cite{shah2014student}, i.e. $\mathbf{x} \sim \mathcal{MVT}( \nu, \boldsymbol{\mu}, \mathbf{K})$, with the dof value $\nu > 2$, mean $\boldsymbol{\mu} \in \mathbb{R}^d$, covariance matrix $\mathbf{K} \in \mathbb{R}^{d \times d}$, its density can be written as
\begin{equation*}
\begin{aligned}
 p(\mathbf{x}) = &
 \frac{\Gamma\left(\frac{\nu+d}{2}\right)}
 {\Gamma\left(\frac{\nu}{2}\right) ((\nu-2)\pi)^{d/2} } |\mathbf{K}|^{-1/2}\\
 &\times
 \left(1 + \frac{(\mathbf{x}-\boldsymbol{\mu})^\top \mathbf{K}^{-1}(\mathbf{x}-\boldsymbol{\mu})}{\nu-2}\right)^{-\frac{\nu+d}{2}}.
\end{aligned}
\end{equation*}

\subsection{The Student's t process}\label{sec:ST process}
The Student's $t$ process (TP)~\cite{shah2014student,solin2015state} is a heavy-tailed, nonparametric distribution over functions and can be viewed as a robust generalisation of the GP~\cite{seeger2004gaussian}. A stochastic process $g$ defined on an input space $\mathcal{X}$ is called a Student's $t$ process with dof $\nu>2$, mean function $\Phi:\mathcal{X}\to\mathbb{R}$, and kernel function $k:\mathcal{X}\times\mathcal{X}\to\mathbb{R}$, denoted by $g \sim \mathcal{TP}(\nu, \Phi, k)$, if for any finite set of inputs $\mathbf{x}=\{x_1,\dots,x_{\eta}\}$,
\begin{equation}\label{eq:ST process}
 (g(x_1),\dots,g(x_{\eta}))^\top \sim \mathcal{MVT}(\nu, \boldsymbol{\mu}, \mathbf{K}),
\end{equation}
where $\boldsymbol{\mu}_\alpha=\Phi(x_\alpha)$, $\mathbf{K}_{\alpha \beta}=k(x_\alpha,x_\beta)$ and $\alpha,\beta = 1,\dots,\eta$. Also, $\mathcal{MVT}(\nu, \boldsymbol{\mu}, \mathbf{K})$ can be written in a Gaussian scale mixture (GSM) form~\cite{solin2015state}.  Let $\gamma^{-1} \sim \Gamma(\nu/2, (\nu-2)/2)$ denote a latent inverse-Gamma scale variable and
\begin{equation}\label{eq:ST GSM form}
    g(\mathbf{x}) \mid \gamma \sim \mathcal{N}\!\left(\phi, \; \gamma\, \mathbf{K}\right),
\end{equation}
then marginalising over $\gamma$ yields the Student’s $t$ process in Eq.~\eqref{eq:ST process}. As the dof parameter $\nu \to \infty$, the  Student's $t$ process converges to a GP:
\begin{equation*}
 \mathcal{TP}(\nu, \Phi, k) \xrightarrow[\nu \to \infty]{} \mathcal{GP}(\Phi, k).
\end{equation*}
Due to its heavy-tailed marginals, the Student's $t$ process assigns higher probability to extreme function values than a GP, leading to improved robustness against outliers.

\subsection{Function-Space Empirical Bayes Regularisation}\label{sec: FSEB_method}
This section describes the FS-EB framework~\cite{rudner2023functionspace,rudner2024finetuning}, which combines parameter-space and function-space regularisation. The FS-EB method considers a supervised learning setting defined by a dataset
\[
\mathcal{D} = \left\{ x^{(n_{\mathcal{D}})}_{\mathcal{D}}, y^{(n_{\mathcal{D}})}_{\mathcal{D}} \right\}_{n_{\mathcal{D}}=1}^{N_{\mathcal{D}}} = \left( \mathbf{x}_{\mathcal{D}}, \mathbf{y}_{\mathcal{D}} \right).
\] 
where the $N_{\mathcal{D}}$ samples are independently and identically distributed. The inputs satisfy
\(
x^{(n_{\mathcal{D}})}_{\mathcal{D}} \in \mathcal{X} \subseteq \mathbb{R}^D
\),
and the corresponding outputs lie in a target space
\(y^{(n_{\mathcal{D}})}_{\mathcal{D}} \in \mathcal{Y}.
\)
For regression tasks, the target space is continuous with
\(
\mathcal{Y} \subseteq \mathbb{R}^L,
\)
whereas for $L$-class classification problems, the targets are represented as $L$-dimensional binary vectors, i.e.,
\(
\mathcal{Y} \subseteq \{0,1\}^L.
\)
The FS-EB defines an auxiliary posterior as an empirical prior, i.e.,
\begin{equation} \label{eq:FSEB aux posterior}
    p(\boldsymbol{\theta} \mid \mathbf{y}_c, \mathbf{x}_c) \propto p(\mathbf{y}_c \mid \mathbf{x}_c, \boldsymbol{\theta})\, p(\boldsymbol{\theta}).
\end{equation}
where $\left( \mathbf{x}_c, \mathbf{y}_c \right)=\left\{ x^{(n_c)}_c, y^{(n_c)}_c \right\}_{n_c=1}^{N_c}$ denotes a set of context points and $p(\boldsymbol{\theta})$ represents the prior distribution over the parameters. Moreover, to formulate the likelihood function $p(\mathbf{y}_c \mid \mathbf{x}_c, \boldsymbol{\theta})$, FS-EB adopts a linear model
\begin{equation}\label{eq: FSEB stochastic model}
z(\mathbf{x}_c) \overset{\cdot}{=} h(\mathbf{x}_c;\boldsymbol{\phi}_0)\boldsymbol{\Psi} + \boldsymbol{\epsilon},
\end{equation}
where $h(\cdot;\boldsymbol{\phi}_0)$ is a feature extractor with parameter $\boldsymbol{\phi}_0$, $\boldsymbol{\Psi} \sim \mathcal{N}(\mathbf{0}, \tau_1 I_1)$ and $\boldsymbol{\epsilon} \sim \mathcal{N}(\mathbf{0}, \tau_2 I_2)$. Also, $I_1$ and $I_2$ are identity matrices, and the positive scalars $\tau_1, \tau_2 \in \mathbb{R}^{+}$ govern the variances of the stochastic components. Give context point set $\mathbf{x}_c$, we have
\begin{equation*}
    p(z(\mathbf{x}_c)\mid\mathbf{x}_c) = \mathcal{N}\big(z(\mathbf{x}_c) ; \mathbf{0},  \mathbf{K}(\mathbf{x}_c, \mathbf{x}_c)\big),
\end{equation*}
with the covariance matrix
\begin{equation}\label{eq: FSEB kernel}
\mathbf{K}(\mathbf{x}_c, \mathbf{x}_c) = \tau_1 h(\mathbf{x}_c;\boldsymbol{\phi}_0) h(\mathbf{x}_c;\boldsymbol{\phi}_0)^\top + \tau_2I_2.
\end{equation}
Viewing this distribution as a likelihood over neural network outputs, FS-EB obtains the following formulation:
\begin{equation}\label{eq: FSEB likelihood}
  p(\mathbf{y}_c^l \mid \mathbf{x}_c, \boldsymbol{\theta}) =
\mathcal{N}\big(\mathbf{y}_c^l; [f(\mathbf{x}_c;\boldsymbol{\theta})]_l,\,
 \mathbf{K}(\mathbf{x}_c, \mathbf{x}_c)\big), 
\end{equation}
where $\mathbf{y}_c = \mathbf{0}$ and $f(\cdot;\boldsymbol{\theta})$ is the neural network parameterised by $\boldsymbol{\theta}$. Also, $[f(\mathbf{x}_c;\boldsymbol{\theta})]_l$ and $\mathbf{y}_c^l$ ($l=1,\cdots,L$) are the $l$-th component of $f(\mathbf{x}_c;\boldsymbol{\theta})$ and $\mathbf{y}_c$, respectively. Then the auxiliary posterior can be expressed as
\begin{equation}\label{eq:FSEB prior}
  p(\boldsymbol{\theta} \mid \mathbf{y}_c, \mathbf{x}_c) \propto p(\boldsymbol{\theta})
\prod_{l=1}^{L}p(\mathbf{y}_c^l \mid \mathbf{x}_c, \boldsymbol{\theta}).  
\end{equation}
Leveraging this auxiliary posterior as an empirical prior, the posterior over parameters given the full dataset is obtained via Bayes' rule:
\begin{equation}\label{eq: FSEB posterior}
p(\boldsymbol{\theta} \mid \mathbf{y}_{\mathcal{D}}, \mathbf{x}_{\mathcal{D}})
\propto p(\mathbf{y}_{\mathcal{D}} \mid \mathbf{x}_{\mathcal{D}}, \boldsymbol{\theta})\,
p(\boldsymbol{\theta} \mid \mathbf{y}_c, \mathbf{x}_c).
\end{equation}
For posterior approximation, \cite{rudner2023functionspace, rudner2024finetuning} introduce techniques grounded in maximum a posteriori (MAP) estimation and VI.

\section{Proposed Method: ST-FS-EB}\label{sec: proposed} 
Following the introduction of the FS-EB framework, this section presents our proposed ST-FS-EB method. In particular, Section~\ref{sec:ST-FS-EB priors} introduces the empirical Student’s t priors over functions, while Section~\ref{sec:ST-FS-EB variational inference} describes the corresponding empirical Bayes variational inference method for posterior approximation.
\subsection{Empirical Student’s t Priors over functions}\label{sec:ST-FS-EB priors}
Analogous to Eq.~\eqref{eq:FSEB aux posterior} in the FSEB framework, we introduce an auxiliary posterior conditioned on the context inputs $\mathbf{x}_c$, given by
\begin{equation*}
    p_{st}(\boldsymbol{\theta} \mid \mathbf{y}_c, \mathbf{x}_c) \propto p_{st}(\mathbf{y}_c \mid \mathbf{x}_c, \boldsymbol{\theta})\, p_{st}(\boldsymbol{\theta}),
\end{equation*}
where both the functional likelihood $p_{st}(\mathbf{y}_c \mid \mathbf{x}_c, \boldsymbol{\theta})$ and the parameter prior $p_{st}(\boldsymbol{\theta})$ are formulated under Student's $t$ assumptions. Let $\theta_i, i=1,\dots,\mathcal{I}$ is the $i$-th component of $\boldsymbol{\theta}$ and $N_{\theta}$ is the number of parameters. We define an independent and identically distributed (i.i.d.) prior over each model parameter as
\begin{equation}\label{eq:distribution of theta}
   p_{st}(\theta_i)=\mathcal{ST}(\theta_i;\nu_{\theta}, \mu_{\theta}, \sigma_{\theta}^2) 
\end{equation}
with zero mean, i.e., $\mu_{\theta} = 0$. Also, following the stochastic modelling strategy in~\eqref{eq: FSEB stochastic model}, we construct the functional likelihood $p_{st}(\mathbf{y}_c \mid \mathbf{x}_c, \boldsymbol{\theta})$ by a stochastic model
\begin{equation*}
z_{st}(\mathbf{x}_c) \overset{\cdot}{=} \sqrt{\gamma}(h(\mathbf{x}_c;\boldsymbol{\phi}_0)\boldsymbol{\Psi} + \boldsymbol{\epsilon}),
\end{equation*}
where $h(\cdot;\boldsymbol{\phi}_0)$, $\boldsymbol{\Psi}$ and  $\boldsymbol{\epsilon}$ have the same meaning as in~\eqref{eq: FSEB stochastic model}, and $\gamma^{-1} \sim \Gamma(\nu_{\theta}/2, (\nu_{\theta}-2)/2)$. Then we have
\begin{equation*}
p(z_{st}(\mathbf{x}_c) \mid \gamma) = \mathcal{N}\!\left(z_{st}(\mathbf{x}_c); \mathbf{0}, \; \gamma\, \mathbf{K}(\mathbf{x}_c, \mathbf{x}_c)\right).    
\end{equation*}
where $\mathbf{K}(\mathbf{x}_c, \mathbf{x}_c)$ can be calculated by \eqref{eq: FSEB kernel}. According to Eq.~\eqref{eq:ST process} and the GSM form in Eq.~\eqref{eq:ST GSM form}, we obtain
\begin{equation*}
p_{st}(z_{st}(\mathbf{x}_c) \mid \mathbf{x}_c) = \mathcal{MVT}\big(z_{st}(\mathbf{x}_c); \nu_{\theta}, \mathbf{0},  \mathbf{K}(\mathbf{x}_c, \mathbf{x}_c)\big).
\end{equation*}
Treating $p_{st}(\mathbf{y}_c^l \mid \mathbf{x}_c, \boldsymbol{\theta})$ as the likelihood model for neural network outputs, we define
\begin{equation}\label{eq: STFSEB likelihood}
  p_{st}(\mathbf{y}_c^l \mid \mathbf{x}_c, \boldsymbol{\theta}) =
\mathcal{MVT}\big( \mathbf{y}_c^l; \nu_{\theta},[f(\mathbf{x}_c;\boldsymbol{\theta})]_l,\,
 \mathbf{K}(\mathbf{x}_c, \mathbf{x}_c)\big) 
\end{equation}
where $[f(\mathbf{x}_c;\boldsymbol{\theta})]_l$ and $\mathbf{y}_c^l=\mathbf{0}$ are the $l$-th component of $f(\mathbf{x}_c;\boldsymbol{\theta})$ and $\mathbf{y}_c$, respectively. Then we have the ST-FS-EB empirical prior
\begin{equation}\label{eq:STFSEB prior}
  p_{st}(\boldsymbol{\theta} \mid \mathbf{y}_c, \mathbf{x}_c) \propto  \prod_{i=1}^{\mathcal{I}}p_{st}(\theta_i)
\prod_{l=1}^{L}p_{st}(\mathbf{y}_c^l \mid \mathbf{x}_c, \boldsymbol{\theta}).  
\end{equation}
\begin{remark}
In the FS-EB framework, both the parameter prior and the functional likelihood are based on Gaussian assumptions. However, existing studies have shown that practical models often exhibit heavy-tailed behaviour in both parameter distributions and output responses~\cite{fortuin2022bayesian}. This empirical evidence motivates the adoption of our Student’s $t$-based prior $ p_{st}(\boldsymbol{\theta} \mid \mathbf{y}_c, \mathbf{x}_c)$, providing a more robust and expressive Bayesian modelling framework.
\end{remark}

\subsection{Empirical Bayes Variational Inference}\label{sec:ST-FS-EB variational inference}
This section formulates posterior inference for the proposed ST-FS-EB prior. In accordance with Eq.~\eqref{eq: FSEB posterior}, the posterior distribution given the full dataset takes the form
\[
p_{st}(\boldsymbol{\theta} \mid \mathbf{y}_{\mathcal{D}}, \mathbf{x}_{\mathcal{D}})
\propto p(\mathbf{y}_{\mathcal{D}} \mid \mathbf{x}_{\mathcal{D}}, \boldsymbol{\theta})\,
p_{st}(\boldsymbol{\theta} \mid \mathbf{y}_c, \mathbf{x}_c).
\]
Due to the intractability of the posterior distribution 
$p_{st}(\boldsymbol{\theta} \mid \mathbf{x}_{\mathcal{D}}, \mathbf{y}_{\mathcal{D}})$, 
we adopt VI to approximate the posterior. Specifically, we seek to minimise the KL divergence between a tractable variational distribution $ q(\boldsymbol{\theta}) $ and the posterior, i.e.,
\[
\min_{q(\boldsymbol{\theta}) \in \mathcal{Q}} D_{\mathrm{KL}}
\big(q(\boldsymbol{\theta}) \,\|\, p_{st}(\boldsymbol{\theta} \mid \mathbf{x}_{\mathcal{D}},
\mathbf{y}_{\mathcal{D}})\big),
\]
where $\mathcal{Q}$ denotes a chosen family of tractable variational distributions. This variational optimisation problem is equivalently formulated as the maximisation of the ELBO:
\begin{align}\label{eq:ELBO}
  \mathcal{L}( \boldsymbol{\theta})
& = \mathbb{E}_{q(\boldsymbol{\theta})}
\big[\log p(\mathbf{y}_{\mathcal{D}} \mid \mathbf{x}_{\mathcal{D}}, \boldsymbol{\theta})\big] \nonumber\\
& \quad - D_{\mathrm{KL}}\big(q(\boldsymbol{\theta}) \,\|\, p_{st}(\boldsymbol{\theta} \mid \mathbf{y}_c, \mathbf{x}_c)\big)  
\end{align}
where the first term corresponds to the expected data log-likelihood under the variational posterior, encouraging accurate data fitting. Besides, the second term acts as a regularisation term that constrains $ q(\boldsymbol{\theta}) $ to remain close to the proposed ST-FS-EB empirical prior $p_{st}(\boldsymbol{\theta} \mid \mathbf{y}_c, \mathbf{x}_c)$.

\paragraph{Variational approximation via MC dropout.}
We employ MC dropout~\cite{gal2016dropout} to construct a tractable variational posterior $q(\boldsymbol{\theta})$. Each stochastic forward pass under a randomly sampled dropout mask yields a realisation $ \boldsymbol{\theta}^{(s)} \sim q(\boldsymbol{\theta}) $, with $ \boldsymbol{\theta}^{(s)} $ denoting the effective parameters associated with the 
$s$-th dropout mask. According to~\eqref{eq:STFSEB prior} and~\eqref{eq:ELBO}, the objective can be approximated as
\begin{align*}
\mathcal{\hat{L}}(\boldsymbol{\theta})
&\approx \frac{1}{S} \sum_{s=1}^{S}
\Bigg[\log p(\mathbf{y}_{\mathcal{D}} \mid \mathbf{x}_{\mathcal{D}}, \boldsymbol{\theta}^{(s)})  \\
&+\sum_{l=1}^{L} \log p_{st}(\mathbf{y}_c^l \mid \mathbf{x}_c,\boldsymbol{\theta}^{(s)}) \Bigg] +\rho\sum_{i=1}^{\mathcal{I}}\mathrm{log}\, p_{st}(\theta_i) 
\end{align*}
where $S$ denotes the number of MC samples and $\rho$ is the MC dropout rate. Then, according to~\eqref{eq:student_t_pdf},~\eqref{eq:distribution of theta},~\eqref{eq: STFSEB likelihood}, we have
\begin{align*}
&\mathcal{\hat{L}}(\boldsymbol{\theta})
\propto\frac{1}{S} \sum_{s=1}^{S}
\Bigg[\log p(\mathbf{y}_{\mathcal{D}} \mid \mathbf{x}_{\mathcal{D}}, \boldsymbol{\theta}^{(s)}) \\
&-\frac{\nu_{\theta}+N_c}{2}\sum_{l=1}^{L} \log \left(1 + \frac{c([f(\mathbf{x}_c;\boldsymbol{\theta}^{(s)})]_l,\mathbf{K}(\mathbf{x}_c, \mathbf{x}_c) )}{\nu_{\theta}-2}\right) \Bigg] \\
&-\frac{\rho(\nu_{\theta}+1)}{2}\sum_{i=1}^{\mathcal{I}} \log \left(1 + \frac{\theta_i^2}{\nu_{\theta}\sigma_\theta^2}\right)   
\end{align*}
where $c(\mathbf{x},\Sigma )\overset{\cdot}{=} \mathbf{x}^\top \Sigma^{-1}\mathbf{x}$ is the squared Mahalanobis distance between $\mathbf{x}$ and the origin under covariance matrix $\Sigma$. The detailed proof is provided in Section~\ref{app:FSVO_MC}. For each epoch of optimisation, the training data $\left( \mathbf{x}_{\mathcal{D}}, \mathbf{y}_{\mathcal{D}} \right)$ is randomly split into a partition of $M$ equally-sized minibatch 
$\left( \mathbf{x}^{(m)}_{\mathcal{B}}, \mathbf{y}^{(m)}_{\mathcal{B}} \right) \sim \left( \mathbf{x}_{\mathcal{D}}, \mathbf{y}_{\mathcal{D}} \right)$, where $m=1,\dots,M$. Let $(\mathbf{x}^{(m)}_c, \mathbf{y}^{(m)}_c) \sim (\mathbf{x}_c, \mathbf{y}_c)$ denote the corresponding randomly sampled context points, we have the loss function at the $m$-th minibatch:
\begin{align} \label{eq:minibatch loss}
    &\mathcal{\hat{L}}^{(m)}(\boldsymbol{\theta})
    \propto \frac{1}{S} \sum_{s=1}^{S}
    \Bigg[\log p(\mathbf{y}^{(m)}_{\mathcal{B}} \mid \mathbf{x}^{(m)}_{\mathcal{B}}, \boldsymbol{\theta}^{(s)}) \nonumber\\
    &-\frac{\nu_{\theta}+N^m_c}{2}\sum_{l=1}^{L} \log \left(1 + \frac{c([f(\mathbf{x}^{(m)}_c;\boldsymbol{\theta}^{(s)})]_l,\mathbf{K}_c^{(m)})}{\nu_{\theta}-2}\right) \Bigg] \nonumber \\
    &-\frac{\rho(\nu_{\theta}+1)}{2M}\sum_{i=1}^{\mathcal{I}} \log \left(1 + \frac{\theta_i^2}{\nu_{\theta}\sigma_\theta^2}\right)   
\end{align}
where $N^m_c$ is the sample size in $(\mathbf{x}^{(m)}_c, \mathbf{y}^{(m)}_c)$ and $\mathbf{K}_c^{(m)}=\mathbf{K}(\mathbf{x}^{(m)}_c, \mathbf{x}^{(m)}_c )$. Also, the last term is scaled by an extra factor $\frac{1}{M}$ as in~\cite{blundell2015weight}.

\begin{algorithm}[t]
\caption{ST-FS-EB training process of each epoch}
\label{alg:st-fseb-simple}
\textbf{Initialisation:} training data $\left( \mathbf{x}_{\mathcal{D}}, \mathbf{y}_{\mathcal{D}} \right)$, model parameters $\boldsymbol{\theta}$, feature extractor $h(\cdot;\boldsymbol{\phi}_0)$, MC dropout rate $\rho$, positive scalars $\tau_1, \tau_2$, dof parameter $\nu_{\theta}$, scale parameter $\sigma_{\theta}$, number of MC samples $S$, number of batches $M$, number of context samples $N^m_c$,

\textbf{for} each minibatch $\left( \mathbf{x}^{(m)}_{\mathcal{B}}, \mathbf{y}^{(m)}_{\mathcal{B}} \right) \sim \left( \mathbf{x}_{\mathcal{D}}, \mathbf{y}_{\mathcal{D}} \right)$:
\begin{enumerate}
\item Sample context points $(\mathbf{x}^{(m)}_c, \mathbf{y}^{(m)}_c)$.
\item Compute loss $\mathcal{\hat{L}}^{(m)}(\boldsymbol{\theta})$ by~\eqref{eq:minibatch loss}.
\item Update parameters $\boldsymbol{\theta}$.
\end{enumerate}
\textbf{end for}
\end{algorithm}

\begin{remark}
The original FS-EB framework assumes Gaussian priors and a Gaussian variational distribution, yielding a closed-form expression for the KL divergence between them. In contrast, our method employs heavy-tailed Student's $t$ priors, which break this conjugacy and therefore preclude a closed-form KL divergence. To accommodate this choice, we leverage MC dropout to implicitly construct the variational posterior, enabling scalable and tractable posterior inference without requiring an explicit parametric form of $q(\boldsymbol{\theta})$.
\end{remark}

\paragraph{Selection of Priors.}
To construct the functional likelihood $p_{st}(\mathbf{y}_c \mid \mathbf{x}_c, \boldsymbol{\theta})$, we choose a feature extractor $h(x;\boldsymbol{\phi}_0)$. In our approach, this extractor is a randomly initialised neural network, which naturally induces inductive biases over functions~\cite{wilson2020bayesian}. We can also consider a pre-trained feature extractor when available, as in~\cite{rudner2023functionspace,rudner2024finetuning}.

\paragraph{Selection of Context Distributions.}
We select context points from an auxiliary dataset $(\mathbf{x}_c, \mathbf{y}_c)$ which is semantically related to but distributionally distinct from the training distribution. For instance, when training on CIFAR-10, we sample context points from CIFAR-100 to provide relevant OOD contextual information.

\begin{table*}[ht]
\centering
\caption{In-distribution prediction performance. \textbf{Best} results are in bold and \underline{second best} are underlined. }
\label{tab:in-dis prediction}
\resizebox{\textwidth}{!}{
\begin{tabular}{l l c c c c c c}
\toprule
Metric & Dataset & ST-FS-EB(our) & FS-EB & GFSVI & MC Dropout & MFVI & MAP \\
\midrule

\multirow{5}{*}{ACC \ $\uparrow$}
& MNIST & \textbf{99.33 ± 0.052} & 99.10 ± 0.067 & 99.17 ± 0.071 & \underline{99.32 ± 0.035} & 99.15 ± 0.095 & 99.09 ± 0.062 \\
& FMNIST & \textbf{92.38 ± 0.194} & 90.49 ± 0.218 & 91.97 ± 0.099 & \underline{92.29 ± 0.189} & 90.28 ± 0.281 & 91.87 ± 0.242 \\
& CIFAR-10 & \underline{86.52 ± 0.395} & 74.42 ± 0.426 & 84.94 ± 0.548 & \textbf{86.54 ± 0.364} & 74.85 ± 0.416 & 77.31 ± 23.656  \\
& PathMNIST & 87.83 ± 1.143 & 86.80 ± 1.338 & \underline{88.32 ± 1.087} & 87.69 ± 0.771 & 87.34 ± 1.253 & \textbf{89.15 ± 2.522}  \\
& OrganAMNIST & \textbf{90.16 $\pm$ 0.635} & 88.56 $\pm$ 0.713 & 87.99 $\pm$ 0.353 & \underline{89.81 $\pm$ 0.560} & 88.55 $\pm$ 0.585 & 86.15 $\pm$ 1.111  \\
\midrule

\multirow{5}{*}{ECE $\downarrow$}
& MNIST & 0.012 ± 0.001 & 0.010 ± 0.002 & \textbf{0.003 ± 0.001} & 0.012 ± 0.001 & \underline{0.008 ± 0.001} & \textbf{0.003 ± 0.001} \\
& FMNIST  & 0.038 ± 0.003 & 0.013 ± 0.003 & \textbf{0.012 ± 0.002} & 0.038 ± 0.004 & \underline{0.011 ± 0.003} & \textbf{0.012 ± 0.003} \\
& CIFAR-10 & \textbf{0.021 ± 0.007} & 0.085 ± 0.004 & \underline{0.023 ± 0.004} & 0.028 ± 0.005 & 0.083 ± 0.003 & \underline{0.023 ± 0.012}  \\
& PathMNIST & 0.049 ± 0.009 & \underline{0.024 ± 0.007} & 0.066 ± 0.008 & 0.055 ± 0.006 & \textbf{0.019 ± 0.003} & 0.068 ± 0.020  \\
& OrganAMNIST & \underline{0.026 $\pm$ 0.007} & 0.038 $\pm$ 0.010 & 0.034 $\pm$ 0.006 & \textbf{0.016 $\pm$ 0.002} & 0.046 $\pm$ 0.003 & 0.051 $\pm$ 0.008  \\
\midrule

\multirow{5}{*}{NLL $\downarrow$}
& MNIST & \underline{0.028 ± 0.001} & 0.034 ± 0.001 & \textbf{0.026 ± 0.002} & 0.029 ± 0.002 & 0.029 ± 0.001 & \textbf{0.026 ± 0.001} \\
& FMNIST  & \textbf{0.226 ± 0.005} & 0.263 ± 0.005 & \underline{0.225 ± 0.002} & 0.227 ± 0.004 & 0.269 ± 0.006 & 0.231 ± 0.005 \\
& CIFAR-10 & \textbf{0.398 ± 0.013} & 0.763 ± 0.009 & 0.466 ± 0.012 & \underline{0.399 ± 0.010} & 0.752 ± 0.008 & 0.642 ± 0.584  \\
& PathMNIST & 0.454 ± 0.069 & \underline{0.405 ± 0.044} & 0.509 ± 0.067 & 0.525 ± 0.113 & \textbf{0.399 ± 0.038} & 0.512 ± 0.121  \\
& OrganAMNIST & \textbf{0.317 $\pm$ 0.012} & 0.364 $\pm$ 0.016 & 0.436 $\pm$ 0.025 & \underline{0.318 $\pm$ 0.016} & 0.360 $\pm$ 0.016 & 0.478 $\pm$ 0.036  \\
\bottomrule
\end{tabular}
}
\end{table*}

\begin{table*}[ht]
\centering
\caption{Out-of-distribution detection performance. \textbf{Best} results are in bold and \underline{second best} are underlined.}
\label{tab:OOD detection}
\resizebox{\textwidth}{!}{
\begin{tabular}{l l c c c c c c}
\toprule
In & OOD &  ST-FS-EB(our) & FS-EB & GFSVI & MC Dropout & MFVI & MAP \\
\midrule

\multirow{3}{*}{MNIST}
& FMNIST & 99.86 ± 0.057 & \textbf{100.00 ± 0.000} & \underline{99.99 ± 0.003} & 98.59 ± 0.351 & 98.60 ± 0.231 & 98.56 ± 0.255 \\
& NotMNIST & \underline{99.96 ± 0.011} & \textbf{99.99 ± 0.003} & \textbf{99.99 ± 0.006} & 96.86 ± 0.286 & 94.06 ± 0.777 & 95.43 ± 0.670 \\
& MNIST-c & 89.38 ± 0.569 & \textbf{90.90 ± 0.665} & \underline{90.61 ± 0.909} & 84.60 ± 0.679 & 82.69 ± 0.943 & 82.30 ± 0.743\\
\midrule

\multirow{2}{*}{FMNIST}
& MNIST & \textbf{99.70 ± 0.170} & \underline{99.63 ± 0.116} & \textbf{99.70 ± 0.116} & 81.22 ± 1.717 & 77.48 ± 2.726 & 78.40 ± 2.310 \\
& NotMNIST & \underline{96.23 ± 0.665} & \textbf{96.97 ± 0.289} & 95.30 ± 0.535 & 80.03 ± 1.611 & 70.57 ± 4.324 & 69.49 ± 1.552 \\
\midrule

\multirow{4}{*}{CIFAR-10}
& SVHN & 84.87 ± 2.269 & 82.53 ± 1.836 & \textbf{88.96 ± 1.524} & \underline{85.32 ± 1.174} & 82.56 ± 2.690 & 82.70 ± 11.716 \\
& CIFAR-10C0 & \underline{55.34 ± 0.429} & 55.08 ± 0.419 & 55.04 ± 0.436 & \textbf{55.40 ± 0.428} & 55.07 ± 0.382 & 54.73 ± 1.724 \\
& CIFAR-10C2 & \underline{64.19 ± 0.784} & 61.71 ± 0.870 & 63.72 ± 0.942 & \textbf{64.49 ± 0.842} & 61.69 ± 0.794 & 62.52 ± 4.470 \\
& CIFAR-10C4 & 71.35 ± 1.225 & 66.27 ± 1.423 & \underline{71.53 ± 1.362} & \textbf{71.77 ± 1.107} & 66.39 ± 1.381 & 68.97 ± 6.767 \\
\midrule
\multirow{1}{*}{PathMNIST}
& BloodMNIST & \textbf{97.29 ± 0.957} & 85.85 ± 10.503 & \underline{96.28 ± 3.110} & 71.24 ± 13.143 & 78.38 ± 7.713 & 81.05 ± 7.870 \\
\midrule

\multirow{1}{*}{OrganAMNIST}
& OrganSMNIST & \underline{87.85 $\pm$ 2.960} & 86.13 $\pm$ 1.015 & \textbf{88.45 $\pm$ 1.601} & 79.80 $\pm$ 0.608 & 77.73 $\pm$ 0.564 & 74.85 $\pm$ 1.058\\

\bottomrule
\end{tabular}
}
\end{table*}

\begin{figure*}
\centering
\includegraphics[width=\linewidth]{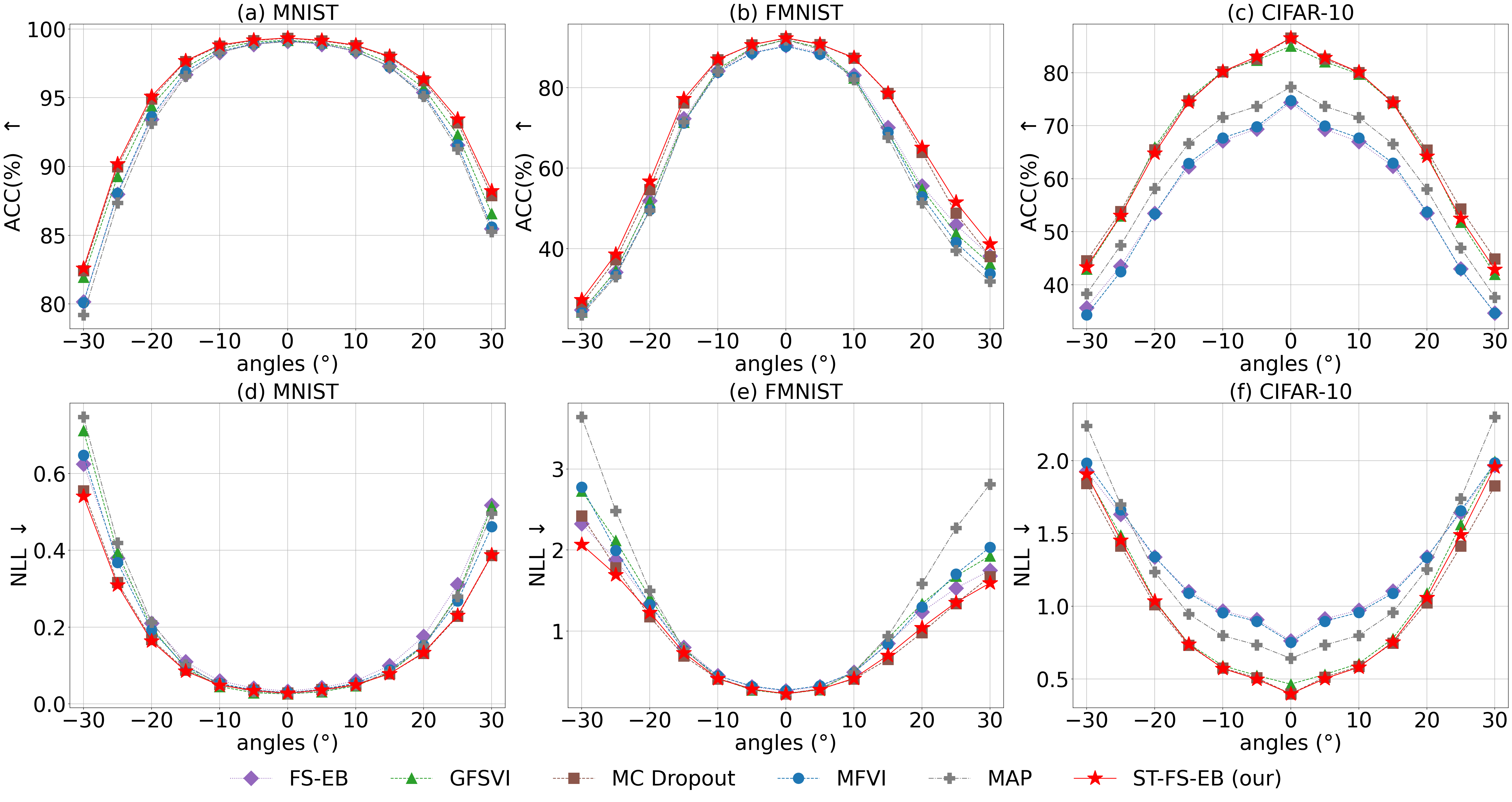}
\caption{Performance under distribution shift induced by image rotations. The top row ((a)--(c)) reports ACC scores, while the bottom row ((d)--(f)) shows NLL. The horizontal axes denote the rotation angle, ranging from $-30^\circ$ to $30^\circ$, and the vertical axis shows the corresponding performance scores.}\label{fig:DS1}
\end{figure*}

\begin{table*}[t]
\centering
\caption{Influence of dof values of Student’s $t$ priors. \textbf{Best} results are in bold and \underline{second best} are underlined.}
\label{tab:effect of dof}
\resizebox{\textwidth}{!}{
\begin{tabular}{l l c c c c c c}
\toprule
IN & Metric/OOD &  2.1 & 3.0 & 5.0 & 10.0 & 20.0 & Gaussian \\
\midrule

\multirow{5}{*}{MNIST}
& ACC & 99.31 $\pm$ 0.031 & 99.31 $\pm$ 0.072 & \underline{99.33 $\pm$ 0.052} & \textbf{99.34 $\pm$ 0.052} & \textbf{99.34 $\pm$ 0.058} & 99.31 $\pm$ 0.044 \\
& NLL & 0.031 $\pm$ 0.001 & 0.030 $\pm$ 0.002 & \underline{0.028 $\pm$ 0.001} & \underline{0.028 $\pm$ 0.002} & \textbf{0.027 $\pm$ 0.001} & 0.031 $\pm$ 0.001 \\
& FMNIST & \underline{99.91 $\pm$ 0.042} & 99.85 $\pm$ 0.035 & 99.86 $\pm$ 0.057 & 99.84 $\pm$ 0.040 & \textbf{99.95 $\pm$ 0.016} & 99.90 $\pm$ 0.036 \\
& NotMNIST & \underline{99.96 $\pm$ 0.011} & 99.95 $\pm$ 0.010 & \underline{99.96 $\pm$ 0.011} & \underline{99.96 $\pm$ 0.011} & \textbf{99.99 $\pm$ 0.004} & 99.95 $\pm$ 0.013 \\
& MNIST-c & \underline{89.53 $\pm$ 0.560} & 89.26 $\pm$ 0.538 & 89.38 $\pm$ 0.569 & 89.25 $\pm$ 0.606 & \textbf{90.22 $\pm$ 0.577} & 89.51 $\pm$ 0.469 \\
\midrule

\multirow{4}{*}{FMNIST}
& ACC & 92.38 $\pm$ 0.194 & \textbf{92.45 $\pm$ 0.211} & \underline{92.40 $\pm$ 0.197} & 92.36 $\pm$ 0.198 & 92.35 $\pm$ 0.117 & 92.25 $\pm$ 0.238 \\
& NLL & 0.226 $\pm$ 0.005 & 0.225 $\pm$ 0.006 & \textbf{0.223 $\pm$ 0.005} & \underline{0.224 $\pm$ 0.006} &\textbf{ 0.223 $\pm$ 0.005} & 0.230 $\pm$ 0.006 \\
& MNIST & \underline{99.70 $\pm$ 0.170} & \textbf{99.83 $\pm$ 0.085} & 91.31 $\pm$ 1.356 & 90.47 $\pm$ 1.766 & 99.05 $\pm$ 0.271 & 97.11 $\pm$ 1.063 \\
& NotMNIST & \underline{96.23 $\pm$ 0.665} & \textbf{96.96 $\pm$ 0.724} & 89.32 $\pm$ 1.844 & 89.07 $\pm$ 1.697 & 95.31 $\pm$ 0.679 & 93.36 $\pm$ 0.630 \\

\midrule
\multirow{6}{*}{CIFAR-10}
& ACC & \textbf{87.17 $\pm$ 0.382} & 86.88 $\pm$ 0.612 & 86.81 $\pm$ 0.680 & \underline{86.95 $\pm$ 0.412} & 86.52 $\pm$ 0.395 & 86.73 $\pm$ 0.349 \\
& NLL & \textbf{0.380 $\pm$ 0.010} & 0.393 $\pm$ 0.015 & 0.393 $\pm$ 0.019 & \underline{0.391 $\pm$ 0.009} & 0.398 $\pm$ 0.013 & 0.402 $\pm$ 0.010 \\
& SVHN & 84.22 $\pm$ 2.104 & 84.40 $\pm$ 2.170 & 83.22 $\pm$ 3.230 & 84.18 $\pm$ 0.913 & \underline{84.87 $\pm$ 2.269} & \textbf{86.51 $\pm$ 2.022} \\
& CIFAR-10C0 & \textbf{55.59 $\pm$ 0.539} & 55.30 $\pm$ 0.447 & 55.50 $\pm$ 0.676 & 55.38 $\pm$ 0.603 & 55.34 $\pm$ 0.429 & \underline{55.58 $\pm$ 0.466} \\
& CIFAR-10C2 & \underline{64.82 $\pm$ 1.007} & 64.34 $\pm$ 0.803 & 64.64 $\pm$ 1.293 & 64.33 $\pm$ 1.175 & 64.19 $\pm$ 0.784 & \textbf{65.02 $\pm$ 0.895} \\
& CIFAR-10C4 & \underline{72.15 $\pm$ 1.471} & 71.87 $\pm$ 1.268 & 71.72 $\pm$ 1.906 & 71.46 $\pm$ 1.454 & 71.35 $\pm$ 1.225 & \textbf{72.55 $\pm$ 1.507} \\

\midrule
\multirow{3}{*}{PathMNIST}
& ACC & 87.57 $\pm$ 1.326 & 87.78 $\pm$ 1.154 & \textbf{88.08 $\pm$ 1.237} & \underline{88.02 $\pm$ 1.012} & 87.83 $\pm$ 1.143 & 88.01 $\pm$ 1.093 \\
& NLL & 0.470 $\pm$ 0.110 & \textbf{0.423 $\pm$ 0.044} & \underline{0.424 $\pm$ 0.050} & 0.458 $\pm$ 0.050 & 0.454 $\pm$ 0.069 & 0.441 $\pm$ 0.046 \\
& BloodMNIST & 97.35 $\pm$ 1.720 & 96.97 $\pm$ 1.317 & \textbf{97.69 $\pm$ 0.973} & \underline{97.43 $\pm$ 1.179} & 97.29 $\pm$ 0.957 & 97.14 $\pm$ 2.531 \\

\midrule
\multirow{3}{*}{OrganAMNIST}
& ACC & \underline{90.50 $\pm$ 0.531} & 90.16 $\pm$ 0.635 & 90.35 $\pm$ 0.273 & 89.65 $\pm$ 0.877 & 90.37 $\pm$ 0.937 & \textbf{90.55 $\pm$ 0.496} \\
& NLL & \underline{0.305 $\pm$ 0.016} & 0.317 $\pm$ 0.012 & 0.314 $\pm$ 0.008 & 0.354 $\pm$ 0.022 & 0.316 $\pm$ 0.028 & \textbf{0.297 $\pm$ 0.011} \\
& OrganSMNIST & 81.90 $\pm$ 0.800 & 87.85 $\pm$ 2.960 & \textbf{89.86 $\pm$ 2.058} & 80.20 $\pm$ 0.948 & 83.16 $\pm$ 2.210 & \underline{87.93 $\pm$ 1.208} \\

\bottomrule
\end{tabular}
}
\end{table*}

\paragraph{Posterior Predictive Distributions.}
At inference time, predictions are formed by averaging the outputs of multiple stochastic forward passes with dropout enabled, thereby approximating the predictive distribution via MC integration under the variational posterior $q(\boldsymbol{\theta})$
\begin{align*}
q(y^* \mid x^*) 
&= \int p\big(y^* \mid f(x^*; \boldsymbol{\theta})\big)\, q(\boldsymbol{\theta})\, d\boldsymbol{\theta} \\
&\approx \frac{1}{\Xi} \sum_{\xi=1}^{\Xi} 
p\big(y^* \mid f(x^*; \boldsymbol{\theta}^{(\xi)})\big),
\end{align*}
where $\boldsymbol{\theta}^{(\xi)} \sim q(\boldsymbol{\theta})$, $\Xi$ denotes the number of MC samples used to estimate the predictive distribution. $x^*$ and $y^*$ denote a test input and its corresponding predicted output, respectively.

\section{Experiments} \label{sec: simulation}
This section presents a comprehensive evaluation of ST-FS-EB, beginning with experimental descriptions—including baselines, setup and implementation details—followed by three sets of experiments: (1) in-distribution prediction and OOD detection (Section~\ref{sec: In and OOD performance}); (2) robustness under distribution shift (Section~\ref{sec:DS1}); (3) an ablation study on the dof values of the Student’s $t$ prior (Section~\ref{sec: dof effect}).

\textbf{Baselines.} 
We benchmark our method against a diverse set of baselines spanning both function-space and parameter-space regularisation paradigms. Specifically, we include two recent FSVI approaches: FS-EB~\cite{rudner2023functionspace} and generalised FSVI (GFSVI)~\cite{cinquin2024regularized}. On the parameter-space side, we consider MC Dropout~\cite{gal2016dropout} and mean-field VI (MFVI)~\cite{blundell2015weight}, along with parameter-space MAP estimation~\cite{bishop2006pattern} as a standard baseline.

\textbf{Setup.} 
We evaluate our method on five in-distribution datasets: MNIST, FashionMNIST (FMNIST), CIFAR-10, PathMNIST, and OrganAMNIST. For OOD detection evaluation, we use
(1)FMNIST, NotMNIST and corrupted MNIST (MNIST-C) for MNIST;  (2) MNIST and NotMNIST for FMNIST; (3)SVHN and corrupted CIFAR-10 with severity levels 0, 2, and 4 (denoted CIFAR-10C0, CIFAR-10C2, CIFAR-10C4) for CIFAR-10; (4) BloodMNIST for PathMNIST; (5) OrganSMNIST for OrganAMNIST. For MNIST-C and CIFAR-10C0/C2/C4, multiple corruption types are considered, and we report the average OOD detection scores across all corruption types as the final evaluation scores. For MNIST, FMNIST and OrganAMNIST, 
we employ a convolutional neural network consisting of two convolutional layers with 32 and 64 filters of size 
$3\times3$, ending with a 128-unit fully connected layer. For CIFAR-10 and PathMNIST, we use a deeper architecture comprising six convolutional layers with 32, 32, 64, 64, 128, and 128 filters (all with $ 3\times3 $ kernels), capped by a dense layer with 128 hidden units. All models are trained using the Adam optimiser, and all images are normalised to the range $ [0, 1]$. For additional details on neural network architectures, hyperparameters and other settings, see Section~\ref{app: Experimental Details}. In-distribution prediction performance is evaluated using classification accuracy (ACC), negative log-likelihood (NLL), and expected calibration error (ECE). For OOD detection, we use the area under the receiver operating characteristic curve (AUROC), with the maximum softmax probability (MSP) employed as the detection score.

\textbf{Implementation.} All experiments are conducted with 10 MC runs, and we report both the mean and standard deviation. For context construction, we use related auxiliary datasets: Kuzushiji-MNIST (KMNIST) for MNIST and FashionMNIST, CIFAR-100 for CIFAR-10, DermaMNIST for PathMNIST and OrganCMNIST for OrganAMNIST. 
Also, the feature extractor $h(\cdot;\boldsymbol{\phi}_0)$  shares the same architecture as the trained neural network, except that the final layer is removed. Moreover, we set $S = 10$ for training and $\Xi = 10$ for inference. 

\subsection{Prediction and OOD Detection Performance}\label{sec: In and OOD performance}
The in-distribution prediction and OOD detection results of our proposed ST-FS-EB, in comparison with all baseline methods, are shown in Table~\ref{tab:in-dis prediction} and Table~\ref{tab:OOD detection}, respectively. As shown in Table~\ref{tab:in-dis prediction}, ST-FS-EB achieves the highest ACC on MNIST, FMNIST, and OrganAMNIST, and remains competitive with the top-performing methods on CIFAR-10 and PathMNIST. In terms of ECE, ST-FS-EB performs competitively on CIFAR-10 and OrganAMNIST, but exhibits higher ECE values than some baselines (e.g., GFSVI and MFVI) on MNIST, FMNIST, and PathMNIST, suggesting its variability in calibration performance across datasets. With respect to NLL, ST-FS-EB consistently achieves the lowest values on FMNIST, CIFAR-10 and OrganAMNIST, while remaining comparable to that of the best-performing baselines. By contrast, Table~\ref{tab:OOD detection} shows that ST-FS-EB consistently achieves the first- or second-best performance across most OOD benchmarks and remains close to the optimal results in the other cases.

Overall, ST-FS-EB achieves a strong balance between predictive accuracy and uncertainty estimation, with robust performance across datasets and metrics. It outperforms FS-EB and GFSVI in in-distribution accuracy while retaining competitive OOD detection. Also, the ST-FS-EB consistently surpasses MC Dropout, MFVI, and MAP on both prediction and OOD benchmarks in most cases.

\subsection{Performance on distribution shifts}\label{sec:DS1}
To investigate the robustness of the proposed method under distribution shift, this experiment introduces controlled rotational transformations to the test data without applying such transformations during training, thereby evaluating the generalisation ability of our proposed method to unseen rotations. Figure~\ref{fig:DS1} shows the ACC and NLL performance of ST-FS-EB and baseline methods under distribution shift on MNIST, FMNIST, and CIFAR-10, while the results for PathMNIST and OrganAMNIST are reported in Figure~\ref{fig:DS2}~(Section~\ref{sec:DS2})). The results demonstrate that the proposed method consistently achieves higher ACC and lower NLL than competing approaches, with only marginal differences compared to MC Dropout in a few scenarios. This highlights the strong robustness of ST-FS-EB under distribution shifts compared with existing baselines, which is critical for safety-sensitive and real-world deployment scenarios.

\subsection{Influence of Degrees of Freedom in the Student-t Prior}\label{sec: dof effect}
This experiment studies the influence of the dof parameter of Student's $t$ priors on the performance of ST-FS-EB. We evaluate a range of dof values $[2.1,3.0,5.0,10.0,20.0]$ and additionally include a Gaussian prior. The experimental results are summarised in Table~\ref{tab:effect of dof}. The optimal dof value exhibits clear dataset dependence for both in-distribution prediction (ACC and NLL) and OOD detection performance. For in-distribution prediction evaluation, no single dof value consistently dominates across datasets. On MNIST and FMNIST, higher dof (dof = 5.0–20.0) generally yield slightly better ACC and NLL performance. On CIFAR-10, dof = 2.1 achieves the optimal results, indicating a preference for heavier-tailed priors. For PathMNIST, dof = 5.0 provides the best performance. In contrast, on OrganAMNIST, the Gaussian prior attains optimal predictive performance. A similar pattern is observed for OOD detection. The best-performing dof configurations vary across benchmarks. For instance,  dof = 20.0 performs the best on MNIST-based OOD tasks, while dof = 3.0 is optimal for FMNIST-based OOD detection. The Gaussian prior remains competitive in CIFAR-10, while dof = 5.0 is generally favourable for PathMNIST and OrganAMNIST. These observations demonstrate that the proposed heavy-tailed Student’s t prior assumption provides meaningful modelling flexibility for real-world data, yielding gains in both predictive accuracy and OOD detection.

\section{Conclusion} \label{sec: conclusion}
This paper proposes ST-FS-EB, a Student’s $t$-based function-space empirical Bayes regularisation framework, which introduces heavy-tailed priors over both parameter and functional distributions. We further derive an MC dropout-based ELBO objective to enable scalable posterior inference. The proposed method is compared with several baseline methods, and experiment results show that ST-FS-EB achieves a superior balance between predictive accuracy and OOD detection performance. Also, the ST-FS-EB exhibits the most robust performance under distribution shifts. Moreover, the optimal dof values of the Student’s $t$ priors are dataset-dependent, and Student’s $t$ priors present better prediction and OOD detection performance than Gaussian priors in most cases.

% References
\bibliography{uai2026-template}

\newpage

\onecolumn

\title{Function-Space Empirical Bayes Regularisation with Student's t Priors\\(Supplementary Material)}
\maketitle

\appendix
\section{The Empirical Bayes Variational Objective VIA MC dropout}\label{app:FSVO_MC}
In this section, we derive the proposed ST-FS-EB variational objective. Starting from Eq.~\eqref{eq:ELBO}, we obtain:
\begin{align*}
\mathcal{L}(\boldsymbol{\theta})
 =& \mathbb{E}_{q(\boldsymbol{\theta})}
\big[\mathrm{log} \ p(\mathbf{y}_{\mathcal{D}} \mid \mathbf{x}_{\mathcal{D}}, \boldsymbol{\theta})\big]- D_{\mathrm{KL}}\big(q(\boldsymbol{\theta}) \,\|\, p_{st}(\boldsymbol{\theta} \mid \mathbf{y}_c, \mathbf{x}_c)\big) \\
=&  \mathbb{E}_{q(\boldsymbol{\theta})}
\big[\mathrm{log} \ p(\mathbf{y}_{\mathcal{D}} \mid \mathbf{x}_{\mathcal{D}}, \boldsymbol{\theta})\big] +\int q(\boldsymbol{\theta})\mathrm{log}\frac{p_{st}(\boldsymbol{\theta} \mid \mathbf{y}_c, \mathbf{x}_c)}{q(\boldsymbol{\theta})} d\boldsymbol{\theta} 
\end{align*}
Substituting the ST-FS-EB prior in Eq.~\eqref{eq:STFSEB prior} into the objective, we obtain
\begin{align*}
\mathcal{L}(\boldsymbol{\theta})
\propto&\mathbb{E}_{q(\boldsymbol{\theta})}
\big[\mathrm{log} \ p(\mathbf{y}_{\mathcal{D}} \mid \mathbf{x}_{\mathcal{D}}, \boldsymbol{\theta})\big] +\int q(\boldsymbol{\theta})\mathrm{log}\frac{\prod_{i=1}^{\mathcal{I}}p_{st}(\theta_i) \prod_{l=1}^{L}p_{st}(\mathbf{y}_c^l \mid \mathbf{x}_c, \boldsymbol{\theta})}{q(\boldsymbol{\theta})} d\boldsymbol{\theta} \\
=&\mathbb{E}_{q(\boldsymbol{\theta})}\big[\mathrm{log} \ p(\mathbf{y}_{\mathcal{D}} \mid \mathbf{x}_{\mathcal{D}}, \boldsymbol{\theta})\big] +
\sum_{l=1}^{L}\mathbb{E}_{q(\boldsymbol{\theta})}\big[\mathrm{log} \big(p_{st}(\mathbf{y}_c^l \mid \mathbf{x}_c, \boldsymbol{\theta})\big)\big] +\sum_{i=1}^{\mathcal{I}}\mathbb{E}_{q(\boldsymbol{\theta})}\big(\mathrm{log}\, p_{st}(\theta_i)\big) -\mathbb{E}_{q(\boldsymbol{\theta})}\big(q(\boldsymbol{\theta})\big)
\end{align*}
To enable a tractable VI formulation, we parameterise the variational distribution $q(\boldsymbol{\theta})$ using MC dropout. Under this approximation, the entropy term $\mathbb{E}_{q(\boldsymbol{\theta})}\big(q(\boldsymbol{\theta})\big)$ becomes a constant~\cite{gal2016dropout}. Let
$\rho$ denotes the dropout rate, we have
\begin{equation*}
\mathcal{L}(\boldsymbol{\theta}) \propto\frac{1}{S} \sum_{s=1}^{S}
\Bigg[\log p(\mathbf{y}_{\mathcal{D}} \mid \mathbf{x}_{\mathcal{D}}, \boldsymbol{\theta}^{(s)}) +
\sum_{l=1}^{L} \log p_{st}(\mathbf{y}_c^l \mid \mathbf{x}_c,\boldsymbol{\theta}^{(s)}) \Bigg] +\rho\sum_{i=1}^{\mathcal{I}}\mathrm{log}\, p_{st}(\theta_i)   
\end{equation*}
where $ \boldsymbol{\theta}^{(s)} \sim q(\boldsymbol{\theta}) $ denotes the effective parameters associated with the $s$-th dropout mask. By Eq.~\eqref{eq: STFSEB likelihood}, we have
\begin{equation*}
\log p_{st}(\mathbf{y}_c^l \mid\mathbf{x}_c,\boldsymbol{\theta}^{(s)})  
=\log \mathcal{MVT}\big(\mathbf{y}_c^l; \nu_{\theta}, [f(\mathbf{x}_c;\boldsymbol{\theta})]_l,\, \mathbf{K}(\mathbf{x}_c, \mathbf{x}_c)\big)  
\end{equation*}
Let $\mathbf{y}_c^l=\mathbf{0}$, then
\begin{align*}
&\log p_{st}(\mathbf{y}_c^l \mid \mathbf{x}_c,\boldsymbol{\theta}^{(s)}) \\
=&\log \mathcal{MVT}\big(\mathbf{y}_c^l; \nu_{\theta},[f(\mathbf{x}_c;\boldsymbol{\theta}^{(s)})]_l,\,
 \mathbf{K}(\mathbf{x}_c, \mathbf{x}_c)\big) \\
=&\log \Bigg[\frac{\Gamma\left(\frac{\nu_{\theta}+N_c}{2}\right)}
 {\Gamma\left(\frac{\nu_{\theta}}{2}\right) ((\nu_{\theta}-2)\pi)^{N_c/2} } |\mathbf{K}(\mathbf{x}_c, \mathbf{x}_c)|^{-1/2}
  \left(1 + \frac{[f(\mathbf{x}_c;\boldsymbol{\theta}^{(s)})]_l^\top \mathbf{K}^{-1}(\mathbf{x}_c, \mathbf{x}_c)[f(\mathbf{x}_c;\boldsymbol{\theta}^{(s)})]_l}{\nu_{\theta}-2}\right)^{-\frac{\nu_{\theta}+N_c}{2}} \Bigg] \\
=&\log \Bigg[\frac{\Gamma\left(\frac{\nu_{\theta}+N_c}{2}\right)}
 {\Gamma\left(\frac{\nu_{\theta}}{2}\right) ((\nu_{\theta}-2)\pi)^{N_c/2} }\Bigg] -\frac{1}{2}\log |\mathbf{K}(\mathbf{x}_c, \mathbf{x}_c)| -\frac{\nu_{\theta}+N_c}{2} \log \left(1 + \frac{[f(\mathbf{x}_c;\boldsymbol{\theta}^{(s)})]_l^\top \mathbf{K}^{-1}(\mathbf{x}_c, \mathbf{x}_c)[f(\mathbf{x}_c;\boldsymbol{\theta}^{(s)})]_l}{\nu_{\theta}-2}\right) \\
\propto& -\frac{\nu_{\theta}+N_c}{2} \log \left(1 + \frac{[f(\mathbf{x}_c;\boldsymbol{\theta}^{(s)})]_l^\top \mathbf{K}^{-1}(\mathbf{x}_c, \mathbf{x}_c)[f(\mathbf{x}_c;\boldsymbol{\theta}^{(s)})]_l}{\nu_{\theta}-2}\right)
\end{align*}
Also, according to \eqref{eq:student_t_pdf} and \eqref{eq:distribution of theta}, $\mathrm{log}\, p_{st}(\theta_i)$ can be calculated by
\begin{align*}
\log p_{st}(\theta_i) 
=&\log \mathcal{ST}(\theta_i; \nu_{\theta}, 0, \sigma_\theta^2) \\
=& \log \Bigg[\frac{\Gamma\left(\frac{\nu_{\theta}+1}{2}\right)}{\Gamma\left(\frac{\nu_{\theta}}{2}\right) \sqrt{\pi\nu_{\theta}\sigma_\theta^2} } \left(1 + \frac{\theta_i^2}{\nu_{\theta}\sigma_\theta^2}\right)^{-\frac{\nu_{\theta}+1}{2}} \Bigg] \\
=& \log \Bigg[\frac{\Gamma\left(\frac{\nu_{\theta}+1}{2}\right)}{\Gamma\left(\frac{\nu_{\theta}}{2}\right) \sqrt{\pi\nu_{\theta}\sigma_\theta^2} }\Bigg] -\frac{\nu_{\theta}+1}{2} \log \left(1 + \frac{\theta_i^2}{\nu_{\theta}\sigma_\theta^2}\right) \\
\propto & -\frac{\nu_{\theta}+1}{2}\log \left(1 + \frac{\theta_i^2}{\nu_{\theta}\sigma_\theta^2}\right)
\end{align*}

Following this, we have our ST-FS-BE VI objective:
\begin{align*}
\mathcal{L}(\boldsymbol{\theta})
\propto&\frac{1}{S} \sum_{s=1}^{S}
\Bigg[\log p(\mathbf{y}_{\mathcal{D}} \mid \mathbf{x}_{\mathcal{D}}, \boldsymbol{\theta}^{(s)}) +
\sum_{l=1}^{L} \log p_{st}(\mathbf{y}_c^l \mid \mathbf{x}_c,\boldsymbol{\theta}^{(s)}) \Bigg] +\rho\sum_{i=1}^{\mathcal{I}}\mathrm{log}\, p_{st}(\theta_i) \\
\propto&\frac{1}{S} \sum_{s=1}^{S}
\Bigg[\log p(\mathbf{y}_{\mathcal{D}} \mid \mathbf{x}_{\mathcal{D}}, \boldsymbol{\theta}^{(s)}) -\frac{\nu_{\theta}+N_c}{2}
\sum_{l=1}^{L} \log \left(1 + \frac{[f(\mathbf{x}_c;\boldsymbol{\theta}^{(s)})]_l^\top \mathbf{K}^{-1}(\mathbf{x}_c, \mathbf{x}_c)[f(\mathbf{x}_c;\boldsymbol{\theta}^{(s)})]_l}{\nu_{\theta}-2}\right) \Bigg] \\
&-\frac{\rho(\nu_{\theta}+1)}{2}\sum_{i=1}^{\mathcal{I}} \log \left(1 + \frac{\theta_i^2}{\nu_{\theta}\sigma_\theta^2}\right)
\end{align*}

\section{Additional Experimental Details} \label{app: Experimental Details}

\subsection{Hyperparameters}\label{app: Hype} 
Table~\ref{tab:hyperparameter_ranges} outlines the search space for the key hyperparameters of our proposed ST-FS-EB method. We perform hyperparameter optimisation using randomised search over this space, running a total of 300 trials for each experiment. The best configuration is chosen based on the lowest NLL achieved on the validation set. All experiments employ early stopping with a maximum of 100 training epochs and a patience of 10 epochs. Additionally, the models are trained across all tasks using the Adam optimiser with a learning rate of $5 \times 10^{-4}$, a numerical stability parameter $\epsilon_1 = 10^{-8}$, momentum parameters $(\beta_1, \beta_2) = (0.9, 0.999)$, and a batch size of 128. Context points are sampled from the context datasets, and unless otherwise specified, their number  $N^m_c$ is fixed at 32. During training, we use $S = 10$ MC samples from $q(\boldsymbol{\theta})$ to approximate the objective; at inference, predictions are averaged over $\Xi = 10$ stochastic forward passes to estimate the predictive distribution.

\begin{table}[htbp]
\centering
\caption{Hyperparameter Ranges}
\label{tab:hyperparameter_ranges}
\begin{tabular}{ll}
\hline
\textbf{Hyperparameters} & \textbf{Range} \\
\hline
$v_\theta$ & $\{2.1, 3.0, 5.0, 10.0, 20.0\}$ \\
$\sigma_\theta$ & $\{10^{\ell} \mid \ell = -6, -5, \dots, 1\}$ \\
$\tau_1$ & $ \{10^{\ell} \mid \ell = -6, -5, \dots, 2\}$ \\
$\tau_2$ & $ \{10^{\ell} \mid \ell = -6, -5, \dots, 2\}$ \\
\hline
\end{tabular}
\end{table}

To ensure a fair and consistent comparison, all baseline methods employ the same optimiser and training protocol—including batch size, learning rate, early stopping criterion and so on. The FS-EB derives the empirical prior from a randomly initialised neural network, consistent with the findings of Wilson et al.~\cite{wilson2020bayesian}. In GFSVI, we use a GP prior with a constant zero-mean function, where the covariance kernel hyperparameters are optimised via mini-batch log marginal likelihood maximisation, as in Milios et al.~\cite{milios2018dirichlet}. Both FS-EB and GFSVI draw their context points from the same context datasets used by ST-FS-EB. Finally, for MC Dropout, MFVI, and MAP methods, the isotropic Gaussian weight prior parameters are chosen using a random search procedure.

\subsection{MNIST, FMNIST and OrganAMNIST}\label{app: mnist}

For MNIST, FMNIST and OrganAMNIST datasets, we adopt a convolutional neural network architecture consisting of two convolutional layers with 32 and 64 filters of size $3 \times 3$, respectively. Each convolutional layer is followed by a ReLU activation and a max-pooling operation. The feature representations are flattened and fed into a fully connected layer with 128 hidden units, before a final linear layer produces the classification outputs. For MC Dropout–based methods, dropout layers are placed after each convolutional block and the fully connected layer, and remain active at inference time to facilitate MC sampling. The dropout rate is set to 0.5 for MNIST and FashionMNIST, and to 0.3 for OrganAMNIST. For MNIST and FashionMNIST, Kuzushiji-MNIST (KMNIST) is used as the context dataset, whereas OrganCMNIST is used as the context dataset for OrganAMNIST. For MNIST and FMNIST, 10\% of the training set is held out as a validation set, while for OrganAMNIST, its own validation set is used.

\subsection{CIFAR-10 and PathMNIST}\label{app: cifar}

For both CIFAR-10 and PathMNIST datasets, we employ a convolutional neural network comprising six convolutional layers with 32, 32, 64, 64, 128, and 128 filters, each of size  $3 \times 3$. ReLU activation functions follow each convolutional layer, while max-pooling layers are applied after the second, fourth, and sixth convolutional layers to progressively reduce the spatial dimensions. The extracted feature maps are flattened and processed by a fully connected layer with 128 hidden units, after which a linear layer produces the final classification outputs. In MC Dropout–based methods, dropout is applied after each convolutional block and the fully connected layer, remaining active at inference to enable MC sampling. The dropout rate is set to 0.1 for both CIFAR-10 and PathMNIST. CIFAR-100 is used as the context dataset for CIFAR-10, while DermaMNIST serves as the context dataset for PathMNIST. For CIFAR-10, 10\% of the training set is held out as a validation set, whereas PathMNIST uses its provided validation split. For both datasets, data augmentation is performed by applying random horizontal flips with a 50\% probability, followed by random cropping with a padding of 4 pixels on all sides.

\section{Additional simulation results}
\subsection{Additional Results under Distribution Shifts}\label{sec:DS2}
\begin{figure*}[!htb]
\centering
\includegraphics[width=0.7\linewidth]{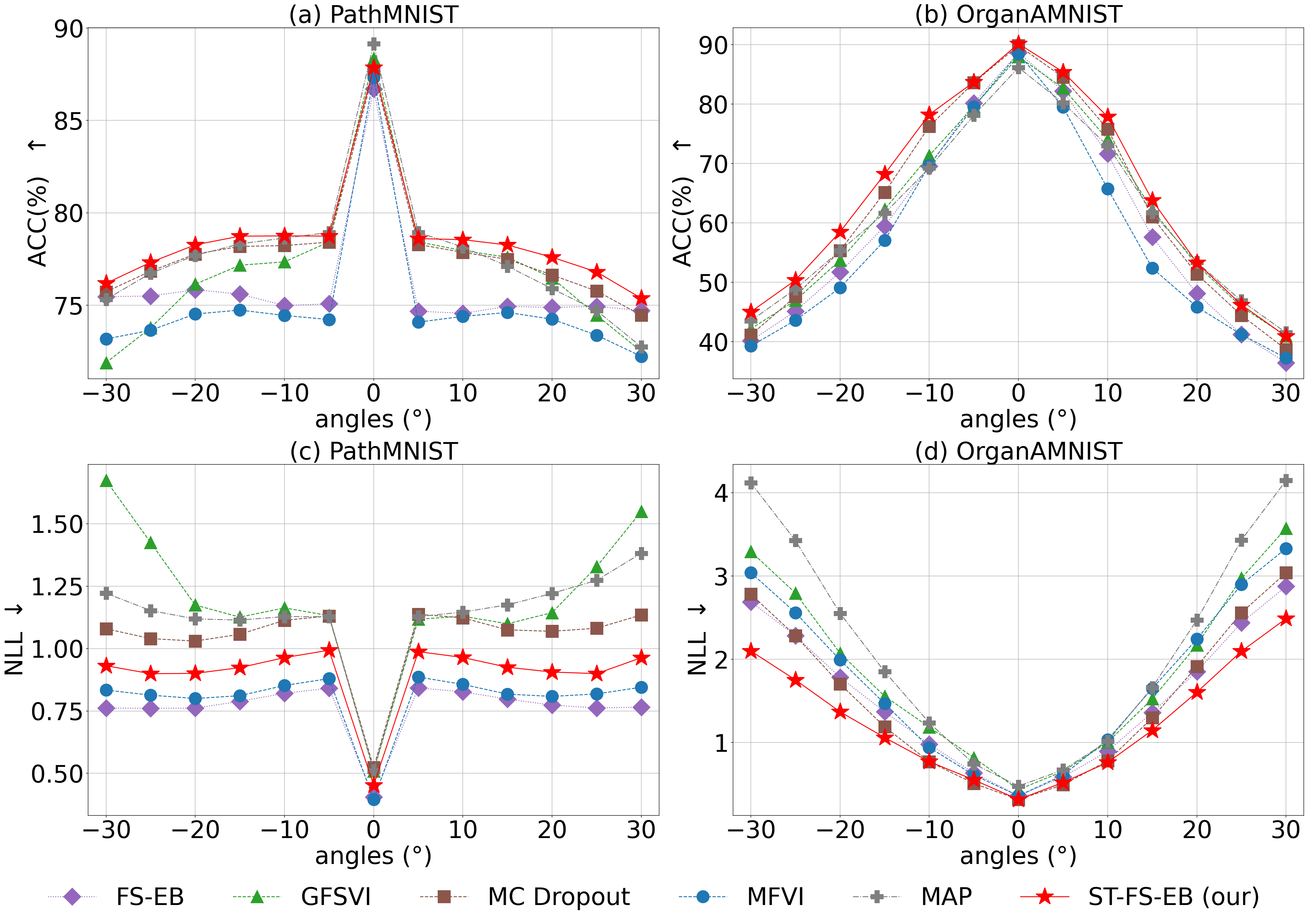}
\caption{Performance under distribution shift induced by image rotations. The top row ((a) (b)) reports ACC scores, while the bottom row ((c) (d)) shows NLL. The horizontal axes denote the rotation angle, ranging from $-30^\circ$ to $30^\circ$, and the vertical axis shows the corresponding performance scores.}\label{fig:DS2}
\end{figure*}

Following the experimental results presented in Section~\ref{sec:DS1}, this section reports additional evaluations under distribution shift. Figure~\ref{fig:DS2} presents the ACC and NLL performance of ST-FS-EB and the baseline methods on PathMNIST and OrganMNIST under the controlled rotational transformations. The results demonstrate that ST-FS-EB consistently achieves superior classification accuracy and lower NLL compared to most competing methods. Although its NLL is higher than that of MFVI and MAP on PathMNIST, ST-FS-EB outperforms the other functional regularisation-based approaches.

\subsection{Influence of the number of context points}
This experiment investigates the impact of the number of context points on model performance. We adopt the same experimental setup as in Section~\ref{sec: In and OOD performance}, and Table~\ref{tab:influence_of_context_number} reports the results across multiple datasets and evaluation metrics. The results indicate that the model is insensitive to the choice of context numbers $\{32, 64, 128\}$, with only marginal variations observed in ACC, NLL and OOD detection performance. This indicates that the proposed method does not depend critically on a specific number of context points and remains robust across different context sizes.
\begin{table*}[ht]
\centering
\caption{Influence of the number of context points. \textbf{Best} results are in bold. OOD detection evaluation: for MNIST, OOD1 = FashionMNIST and OOD2 = NotMNIST; for FashionMNIST, OOD1 = MNIST and OOD2 = NotMNIST; for CIFAR-10, OOD1 = SVHN and OOD2 = CIFAR-10C0; for PathMNIST, OOD1=BloodMNIST; for OrganAMNIST, OOD1=OrganSMNIST.}
\label{tab:influence_of_context_number}
\resizebox{0.85\textwidth}{!}{
\begin{tabular}{l c c c c c c}
\toprule
Metric & Context number & MNIST & FMNIST & CIFAR-10 & PathMNIST & OrganAMNIST \\
\midrule

\multirow{3}{*}{ACC}
& 32  & \textbf{99.33 ± 0.052} & 92.38 ± 0.194 & 86.52 ± 0.395 & \textbf{87.83 ± 1.143} & 90.16 ± 0.635 \\
& 64  & 99.31 ± 0.035 & \textbf{92.42 ± 0.268} & 86.57 ± 0.429 & 87.02 ± 1.346 & 90.60 ± 0.457 \\
& 128 & \textbf{99.33 ± 0.047} & 92.35 ± 0.218 & \textbf{86.66 ± 0.245} & 87.78 ± 1.353 & \textbf{90.65 ± 0.386} \\
\midrule

\multirow{3}{*}{NLL}
& 32  & \textbf{0.028 ± 0.001} & \textbf{0.226 ± 0.005} & \textbf{0.398 ± 0.013} & 0.454 ± 0.069 & 0.317 ± 0.012 \\
& 64  & 0.029 ± 0.002 & 0.229 ± 0.007 & \textbf{0.398 ± 0.012} & 0.448 ± 0.066 & 0.297 ± 0.017 \\
& 128 & 0.030 ± 0.002 & 0.230 ± 0.006 & 0.399 ± 0.007 & \textbf{0.437 ± 0.052} & \textbf{0.291 ± 0.013} \\
\midrule

\multirow{3}{*}{OOD1}
& 32  & 99.86 ± 0.057 & \textbf{99.86 ± 0.057} & 84.87 ± 2.269 & \textbf{97.29 ± 0.957} & \textbf{87.85 ± 2.960} \\
& 64  & 99.92 ± 0.022 & 99.85 ± 0.066 & 86.41 ± 2.526 & 96.79 ± 1.176 & 83.15 ± 1.200 \\
& 128 & \textbf{99.98 ± 0.005} & 99.69 ± 0.141 & \textbf{86.60 ± 1.963} & 97.25 ± 1.896 & 84.50 ± 1.004 \\
\midrule

\multirow{3}{*}{OOD2}
& 32  & 99.96 ± 0.011 & 96.23 ± 0.665 & 71.35 ± 1.225 & -- & -- \\
& 64  & 99.97 ± 0.013 & \textbf{97.17 ± 0.616} & 72.01 ± 1.533 & -- & -- \\
& 128 & \textbf{99.99 ± 0.003} & 96.45 ± 0.676 & \textbf{72.64 ± 1.271} & -- & -- \\
\bottomrule
\end{tabular}
}
\end{table*}

\subsection{Influence of training context distributions}
This experiment investigates the impact of the context distributions on model performance. We adopt the same experimental setup as in Section~\ref{sec: In and OOD performance} and consider two types of context points: (i) samples drawn from the training batches (TRAIN), and (ii) samples drawn from auxiliary context datasets $\mathbf{x}_c$, as defined in Section~\ref{sec: simulation} and Section~\ref{app: Experimental Details}. Table~\ref{tab:Influence_of_Context_Points} reports the results across multiple datasets and evaluation metrics. When context points are sampled from the training dataset (TRAIN), the model consistently achieves higher ACC scores. In contrast, using auxiliary datasets leads to a slight degradation in ACC. However, this reduction is accompanied by improved uncertainty qualification, reflected by lower ECE and NLL, as well as substantially better OOD detection performance. These results indicate that auxiliary context datasets enable a more favourable trade-off between predictive accuracy and uncertainty quantification.

\begin{table*}[ht]
\centering
\caption{Influence of the context distributions. \textbf{Best} results are in bold. OOD detection evaluation: for MNIST, OOD1 = FashionMNIST and OOD2 = NotMNIST; for FashionMNIST, OOD1 = MNIST and OOD2 = NotMNIST; for CIFAR-10, OOD1 = SVHN and OOD2 = CIFAR-10C0; for PathMNIST, OOD1=BloodMNIST; for OrganAMNIST, OOD1=OrganSMNIST.}
\label{tab:Influence_of_Context_Points}
\resizebox{0.9\textwidth}{!}{
\begin{tabular}{l l c c c c c}
\toprule
Dataset & Setting & ACC & ECE & NLL & OOD1 & OOD2 \\
\midrule
MNIST   & TRAIN   & \textbf{99.37 ± 0.048} &  0.015 ± 0.001   &  0.032 ± 0.001   &  99.22 ± 0.156    &  97.99 ± 0.460    \\
        & $\mathbf{x}_c$   & 99.33 ± 0.052         &  \textbf{0.012 ± 0.001} &  \textbf{0.028 ± 0.001} &  \textbf{99.86 ± 0.057} &  \textbf{99.96 ± 0.011} \\
\midrule
FMNIST  & TRAIN   & \textbf{92.43 ± 0.237} &  0.039 ± 0.004   &  \textbf{0.226 ± 0.005} &  85.56 ± 2.097     &  81.62 ± 2.170     \\
        & $\mathbf{x}_c$   & 92.38 ± 0.194         &  \textbf{0.038 ± 0.003} &  \textbf{0.226 ± 0.005} &  \textbf{99.70 ± 0.170} &  \textbf{96.23 ± 0.665} \\
\midrule
CIFAR-10 & TRAIN   & \textbf{86.53 ± 0.445} &  0.027 ± 0.006   &  0.400 ± 0.013   &  \textbf{84.97 ± 2.594} &  \textbf{71.95 ± 1.478} \\
        & $\mathbf{x}_c$   & 86.52 ± 0.395         &  \textbf{0.021 ± 0.007} &  \textbf{0.398 ± 0.013} &  84.87 ± 2.269    &  71.35 ± 1.225   \\
\midrule
PathMNIST    & TRAIN   & \textbf{88.18 ± 1.160} &  0.053 ± 0.009    &  0.492 ± 0.156    &  76.23 ± 22.597     &  --    \\
        & $\mathbf{x}_c$   & 87.83 ± 1.143         &  \textbf{0.049 ± 0.009} &  \textbf{0.454 ± 0.069} &  \textbf{97.29 ± 0.957}   &  --   \\
\midrule
OrganAMNIST     & TRAIN   & \textbf{91.21 ± 0.382} & \textbf{0.016 ± 0.002} & \textbf{0.276 ± 0.008} &  82.13 ± 0.469 & --  \\
        & $\mathbf{x}_c$   & 90.16 ± 0.635 & 0.026 ± 0.007 & 0.317 ± 0.012 & \textbf{87.85 ± 2.960} & --  \\ 
\bottomrule
\end{tabular}
}
\end{table*}

\subsection{Training Efficiency}
This experiment investigates the training efficiency of the proposed method. We adopt the same experimental setup as in Section~\ref{sec: In and OOD performance}. Table~\ref{tab:Training time and memory} shows the average per-epoch training time and memory consumption of ST-FS-EB in comparison with various baselines across multiple datasets. ST-FS-EB incurs a higher computational cost than traditional weight-space regularisation methods, such as MFVI and MAP, reflecting the additional overhead introduced by functional regularisation. However, compared to other function-space approaches (FS-EB and GFSVI), ST-FS-EB is substantially more efficient, requiring less training time and lower memory usage across all benchmarks. This efficiency advantage arises from two factors: 1) ST-FS-EB avoids linearisation-based approximations, thereby eliminating the computational burden of Jacobian evaluations; 2) the use of MC dropout induces sparsity in the network structure, which reduces both computational and memory costs during training.

\begin{table*}[ht]
\centering
\caption{Training time and memory.}
\label{tab:Training time and memory}
\resizebox{0.85\textwidth}{!}{
\begin{tabular}{l l c c c c c c}
\toprule
Metric & Dataset & ST-FS-EB(our) & FS-EB & GFSVI & MC Dropout & MFVI & MAP \\
\midrule

\multirow{4}{*}{Time (s/epoch) $\downarrow$}
& MNIST   & 18.52  & 26.96& 40.44 &11.24 &18.13  & 17.86 \\
& FMNIST   &18.10  &26.63 & 40.76 &11.17 &17.87  &17.62\\
& CIFAR-10  &43.89  &58.69 & 93.01 &32.09 &44.39  & 18.45\\
& PathMNIST     & 74.34 &97.36 & 136.86 &52.39 &71.82  &45.63 \\
& OrganAMNIST     &20.42  & 25.33 &35.62  &16.04 &22.02  &15.24 \\
\midrule

\multirow{4}{*}{Memory (MB) $\downarrow$}
& MNIST   & 536.14 &808.39 &1055.66  &437.84 &633.32  & 67.09 \\
& FMNIST   & 536.14 & 808.39&1055.66  &437.84 &633.32  &67.09\\
& CIFAR-10  & 1590.82 &2060.54 & 2605.42 &1275.60 &1714.07  &112.83 \\
& PathMNIST     &1225.31  &1745.45 &2098.75  &983.55 &1479.45  &93.06 \\
& OrganAMNIST   &535.58  &807.85 &1096.22  &437.28 &632.76  &67.09 \\
\bottomrule
\end{tabular}
}
\end{table*}

\end{document}